\documentclass[runningheads]{llncs}
\usepackage{graphicx}
\usepackage{multirow}
\usepackage{bbm,bm}
\usepackage{tikz}
\usepackage{comment}
\usepackage{amsmath,amssymb} 
\usepackage{color}

\usepackage[accsupp]{axessibility}  

\begin{document}
\pagestyle{headings}
\mainmatter
\def\ECCVSubNumber{3298}  

\title{Uncertainty Inspired Underwater Image Enhancement} 


\titlerunning{Uncertainty Inspired Underwater Image Enhancement}
%
\author{Zhenqi Fu\inst{1} \and
Wu Wang\inst{1} \and
Yue Huang\inst{1}\and
Xinghao Ding\inst{1}\thanks{Corresponding author.}\and
Kai-Kuang Ma\inst{2}
}
\authorrunning{Zhenqi Fu, Wu Wang et al.}
%
\institute{Xiamen University, Fujian, 361005, China\\
\email{\{fuzhenqi,23320170155546\}@stu.xmu.edu.cn}, \email{\{yhuang2010,dxh\}@xmu.edu.cn}\\ \and
Nanyang Technological University, 639798, Singapore\\
\email{ekkma@ntu.edu.sg}}
\maketitle

\begin{abstract}
A main challenge faced in the deep learning-based Underwater Image Enhancement (UIE) is that the ground truth high-quality image is unavailable. Most of the existing methods first generate approximate reference maps and then train an enhancement network with certainty. This kind of method fails to handle the ambiguity of the reference map. In this paper, we resolve UIE into distribution estimation and consensus process. We present a novel probabilistic network to learn the enhancement distribution of degraded underwater images. Specifically, we combine conditional variational autoencoder with adaptive instance normalization to construct the enhancement distribution. After that, we adopt a consensus process to predict a deterministic result based on a set of samples from the distribution. By learning the enhancement distribution, our method can cope with the bias introduced in the reference map labeling to some extent. Additionally, the consensus process is useful to capture a robust and stable result. We examined the proposed method on two widely used real-world underwater image enhancement datasets. Experimental results demonstrate that our approach enables sampling possible enhancement predictions. Meanwhile, the consensus estimate yields competitive performance compared with state-of-the-art UIE methods. Code available at https://github.com/zhenqifu/PUIE-Net. 

\end{abstract}

\section{Introduction}
Underwater images suffer from degradation due to the poor and complex lighting conditions in the water. The degradation of underwater images is mainly rooted in the wavelength-dependent light scattering and absorption, which reduces visibility, decreases contrast, and introduces unpleasant color casts. It is important and necessary to develop Underwater Image Enhancement (UIE) methods to adjust the degraded underwater signal so that the results are more suitable for display or further analysis. In the past few years, deep learning-based UIE approaches have enabled tremendous progress. Commonly, this kind of method adopts corresponding pairs of clean and distorted images to learn a mapping between two quality levels~\cite{1,2,3,4}.

UIE is an important low-level vision task. For underwater scenes with adverse visual conditions, it is difficult and impractical to capture the clean image to train a deep neural network directly. This is because the degradation of underwater images is non-reversible and very complex. To solve this problem, previous methods propose to generate approximate supervisors to train a deep neural network. For example, in~\cite{1} the authors applied several state-of-the-art algorithms to generate a set of potential reference images and manually select the best one as the ground truth. Benefiting from the large-scale dataset constructed in previous works, deep learning-based methods have made profound progress in learning the mapping from a degraded underwater image to the corresponding high-quality reference image~\cite{33}. Nonetheless, considering the progress of UIE under this pipeline, we would like to argue that this kind of method fails to capture the uncertainties in labeling the ground truth. 

Although the reference image achieves high visual quality in existing UIE datasets, it is generated through an approximate approach (e.g., using exiting UIE algorithms~\cite{1}) and may be influenced by various factors, such as human-specific preference during subjective selection and different algorithm parameters. In this case, UIE suffers from uncertainty issues. For ambiguous labels, directly learning a mapping between the degraded underwater image and corresponding reference is inappropriate. There are many potential solutions for the same degraded underwater image because we cannot confidently know what a true clean image looks like. Nonetheless, most of the existing deep learning-based methods treat UIE as a point estimation problem. As a result, they have to make a compromise between possible solutions because they are following a deterministic learning pipeline.

In this paper, we propose the first probabilistic network for UIE, termed PUIE-Net. Instead of directly generating a single prediction (i.e., the point estimation), we are interested in how the network produces multiple results (i.e., the distribution estimation) so that the network can handle the uncertainty issue in UIE. Furthermore, once the distribution is estimated, we can perform a consensus process to capture a deterministic result based on a set of estimations. In this paper, we introduce two consensus processes to predict the final result named: Monte Carlo likelihood estimation (MC)~\cite{5} and Maximum Probability estimation (MP). Specifically, MC is calculated by taking the average of the likelihoods. While the sample with maximum probability is regarded as the final result in MP. 

The proposed network structure is based on probabilistic adaptive instance normalization (PAdaIN), which combines conditional variational autoencoder (CVAE)~\cite{7} with adaptive instance normalization (AdaIN)~\cite{6} to construct the enhancement distribution. PAdaIN is motivated by AdaIN that is originally designed for style transfer. We extend AdaIN into PAdaIN via drawing style inputs from two posterior distributions constructed by conditional variational autoencoders. PAdaIN aims to transform the global enhancement statistics of input features. Therefore, diverse predictions can be achieved by sampling different enhancement attributes from the latent space. The whole pipeline of our method is trained following the standard training procedure of the CVAE. We conduct extensive experiments on two real-world UIE datasets to validate the effectiveness of our method. 

Our main contributions are summarized as follows:

\begin{itemize}
	
	\item We resolve UIE into distribution estimation and consensus process to handle the uncertainty issue in labeling the ground truth. 
	
	\item We propose the first probabilistic network for UIE, which learns to approximate the posterior over meaningful appearance. Specifically,  the enhancement distribution is constructed based on conditional variational autoencoder and adaptive instance normalization.
	
	\item  We show that our method can generate diverse potential solutions. Besides, by inferring the consensus prediction based on a set of samples, our method achieves promising performance compared with state-of-the-art methods on two UIE datasets. 
	
\end{itemize}

\section{Related Work}

\subsection{Underwater Image Enhancement}

According to the means of the modeling imaging process, the existing UIE methods can be roughly categorized into the following three types.

The first category is model-free methods, which enhance underwater images without considering the degradation process. Traditional contrast limited adaptive histogram equalization (CLAHE)~\cite{8}, white balance (WB)~\cite{9}, and Retinex~\cite{10} belong to this category. In~\cite{11}, the authors proposed a fusion-based method for UIE, where the inputs and weight measures are derived only from the degraded image. An improvement version of~\cite{11} is presented in~\cite{12}, which adopts a white balancing technique and a novel multi-scale fusion strategy to further promote the enhancement performance. Fu et al.~\cite{13} proposed a retinex-based UIE approach to enhance a single underwater image. Gao et al.~\cite{14} presented a teleost fish retina-guided underwater image enhancement approach to deal with the problems of nonuniform color shift and content blurring. Other relevant works can be found in~\cite{15,16}. 

The second category is prior-based methods, which enhance underwater images using physical imaging models and focus on accurately estimating the parameters of the defined physical model. Chiang et al.~\cite{17} proposed to enhance underwater images via a dehazing algorithm. Galdran et al.~\cite{19} proposed a variant of the Dark Channel Prior (DCP)~\cite{18} that uses red channel information to estimate the depth map of underwater images. Li et al.~\cite{20} proposed an underwater image dehazing algorithm and a contrast enhancement method based on a minimum information loss and histogram distribution prior. Berman et al.~\cite{21} took into account multiple spectral profiles of different water types, in which the authors additionally estimated two global parameters, i.e., the attenuation ratios of the blue-red and blue-green color channels. Akkaynak et al.~\cite{22} developed a UIE method named Sea-thru based on a revised physical imaging model. Sea-thru takes RGBD images as the input and it first estimates backscatter using the darkest pixels and their known range map. Then it calculates the attenuation coefficient based on an estimation of the spatially varying illuminant. Other relevant works of prior-based UIE methods can be found in~\cite{23,24,25,26,61}. 

The third category is deep learning-based methods that automatically extract representations and learn an enhancement mapping based on numerous paired/unpaired training data. Li et al.~\cite{27} first proposed a generative adversarial network to generate synthetic underwater images in an unsupervised pipeline. Then the authors trained an enhancement network using these synthetic data. Li et al.~\cite{28} proposed a weakly supervised underwater image enhancement method that relaxes the need for paired data. Guo et al.~\cite{29} enhanced degraded underwater images using a multi-scale dense generative adversarial network.  Li et al.~\cite{23} introduced a lightweight UIE model based on the underwater scene prior. Li et al.~\cite{1} constructed a large scale real-world UIE dataset. The reference image is generated by 12 existing UIE methods. Besides, based on this dataset, the authors proposed a gated fusion network for enhancing underwater images. Jamadandi et al.~\cite{31} proposed to enhance underwater images by augmenting the network with wavelet corrected transformations. To deal with the challenge of underwater image degradation diversity, Uplavikar et al.~\cite{32} trained a new deep neural network to learn the domain agnostic features for a given degraded underwater image, where the domain is the Jerlov water type of the image. Li et al.~\cite{33} presented an underwater image enhancement network called Ucolor by medium transmission-guided multi-color space embedding. Kar et al.~\cite{34} presented a zero-shot underwater and hazing image restoration method by leveraging a theoretically deduced property of degradation through the physical model. More relevant works of learning-based UIE methods can be found in~\cite{35,36,37,38,39,40,41,62}. 

\subsection{VAE based Deep Probabilistic Model} 

Variational autoencoder (VAE) and its conditional counterpart (CVAE)~\cite{5,7} have been widely used in various computer vision tasks. Rather than building an encoder that outputs a single value to describe each latent state attribute, VAE formulates the encoder to describe a probability distribution for each latent attribute. To train a VAE, a regularizer and a reconstruction loss are needed to penalize the disagreement of the posterior and prior distributions of the latent representation. 

Despite not being explored to model ambiguities for underwater image enhancement, VAEs and CVAEs are utilized to sample diverse results from constructed posteriors. For example, in~\cite{42}, the authors use VAEs to model the image background for salient object detection. In~\cite{43}, the authors apply VAEs for learning motion sequence generation. In~\cite{44}, a probabilistic U-Net based on VAEs is proposed to learn a conditional distribution of medical image segmentation. \cite{45} and~\cite{46} improve the diversity of samples in~\cite{44} via adopting a hierarchy of latent variables. In~\cite{47}, a contrastive VAE is introduced which combines the benefits of contrastive learning with the power of VAEs to identify and enhance salient latent features.  In~\cite{49}, a VAE based denoising approach is developed by predicting a whole distribution of denoised images. In~\cite{48}, the authors employ VAEs to predict multiple deprojected instances for images/videos collapsed along a dimension. In~\cite{50}, the authors employ VAEs for RGB-D saliency detection by learning from the data labeling process. Other relevant works about VAEs for diverse solution sampling can be found in~\cite{51,52,53}

\section{Method}

In this section, we will introduce PUIE-Net in detail. PUIE-Net is based on PAdaIN that learns meaningful distributions of UIE. It is the first time that employs a probabilistic network to solve the UIE problem.

\subsection{Motivation}
\subsubsection{Ambiguities for UIE}
The main idea of PUIE-Net is to introduce ambiguities for UIE. This is because the true clean image is unavailable and there is a degree of uncertainties in recording approximate labels. Existing deterministic learning-guided methods fail to capture such uncertainty and have to make a compromise between possible results. We consider that as we cannot confidently know which of the possible clean image has given rise to the distorted underwater image at hand, estimating the distribution of possible interpretations may be an advisable solution. We use an implicit variable ${\bf{z}}$ to express the uncertainty. Here, ${\bf{z}}$ can be interpreted as human subjective preferences or camera/algorithm parameters in capturing the ground truth. Let ${\bf{x}}$ and ${\bf{y}}$ refer to the corrupted observation and the clean image, respectively. UIE under a probabilistic framework can be formulated as:
\begin{equation}
p\left( {y\left| {\bf{x}} \right.} \right) \approx \frac{1}{S}\sum\limits_{s = 1}^S {p\left( {{\bf{y}}\left| {{{\bf{z}}^{\left( s \right)}},{\bf{x}}} \right.} \right),{{\bf{z}}^{\left( s \right)}} \sim p\left( {{\bf{z}}\left| {\bf{x}} \right.} \right)} 
\end{equation}
where $p\left( {\bf{z}\left| \bf{x} \right.} \right)$ denotes the distribution of uncertainty. $S$ represents the number of samples. Equation (1) not only allows us to generate multiple enhancement predictions but also gives a straightforward way to calculate a deterministic result, i.e., the MC estimation~\cite{5} that draws samples using the prior network and takes the average of the likelihoods. Apart from MC, the Maximum Probability estimation (MP) is also considered in this paper. MP takes the enhancement sample with the maximum probability as the final result. Mathematically, MP can be expressed as: 
\begin{equation}
p\left( {y\left| {\bf{x}} \right.} \right) \approx p\left( {{\bf{y}}\left| {{{\bf{z}}_{\max }},{\bf{x}}} \right.} \right),{{\bf{z}}_{\max }} \sim p\left( {{\bf{z}}\left| {\bf{x}} \right.} \right)
\end{equation}
where ${\bf{z}}_{\max }$ denotes the sample with the maximum probability.

\subsubsection{PAdaIN for Learning Appearance Distribution}
Since the goal of UIE is to adjust the image appearance such as colors and contrasts, rather than the content, it is important to capture such information during the enhancement. Here, we adopt a modified AdaIN~\cite{6} to capture such features. AdaIN is originally developed for style transfer, which can be expressed as:
\begin{equation}
{\rm{AdaIN}} \left( {\bf{x, y}} \right) = {\bm{\sigma}\bf{(y)}}\left( {\frac{{{\bf{x}} - \bm{\mu}\left( {\bf{x}} \right)}}{{\bm{\sigma}\left( {\bf{x}} \right)}}} \right) + {\bm{\mu}\bf{(y)}}
\end{equation}
where $\bf{x}$ denotes the features of the content image and $\bf{y}$ denotes the features of the style image. $\bm{\mu}$ and $\bm{\sigma}$ refer to the mean and standard deviation operations, respectively. AdaIN changes the appearance by adjusting the mean and standard deviation of features. We observe that UIE falls into the AdaIN paradigm. However, AdaIN relies on the known content and style images, it cannot be directly leveraged for UIE. To handle this problem, we propose PAdaIN, which can be formulated as:  
\begin{equation}
{\rm{PAdaIN}} \left( {\bf{x}} \right) = {\bm{b}}\left( {\frac{{{\bf{x}} - \bm{\mu}\left( {\bf{x}} \right)}}{{\bm{\sigma}\left( {\bf{x}} \right)}}} \right) + {\bm{a}}
\end{equation}
where $\bm{b}$ and $\bm{a}$ are two random samples from the posterior distributions of the mean and standard deviation, respectively. Specifically, the posterior distributions can be learned with CVAEs, which will be detailed in the next subsection. Note that the proposed PAdaIN can be regarded as a generalized AdaIN. 

\subsection{Network Structure}

\begin{figure}[!t]
	\centerline{\includegraphics[scale=0.45]{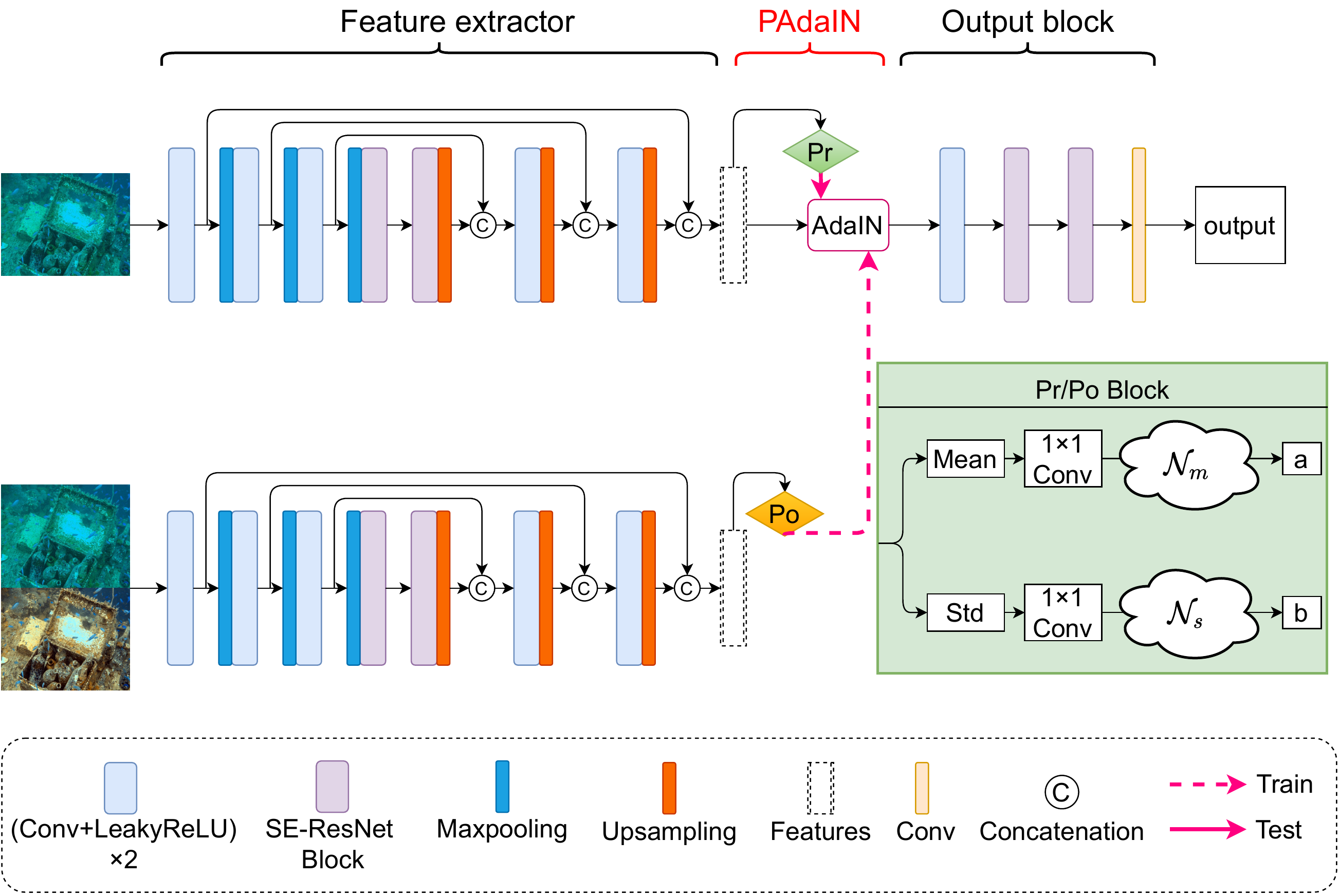}}
	\caption{\textbf{The network architecture of PUIE-Net.} The feature extractor is based on U-Net~\cite{54} that maps the input to representations. After the feature extractor is the PAdaIN module, which transforms the enhancement statistics of received deterministic features. \textbf{In the training phase,} features from the bottom branch are used to calculate the posterior distribution. Random samples from the posterior distributions are injected into AdaIN to transform the enhancement representation. \textbf{In the testing phase,} taking a single degraded image as the input, random samples from the Pr block are employed to generate the predictions.}
\end{figure}

The whole pipeline of PUIE-Net is illustrated in Fig. 1. The proposed network structure contains two branches. Both branches include a U-Net~\cite{54} based feature extractor. Specifically, the top branch aims to estimate the prior distribution of a single raw underwater image. While the goal of the bottom branch is to construct posterior distributions of UIE, it takes the raw underwater image and corresponding reference image as the input. In PUIE-Net, we simply modify the U-Net by adding SE-ResNet blocks~\cite{55}. This is useful to improve the network capacity of enhancement representations.

The core component of PUIE-Net is the PAdaIN after the feature extractor to encode the ambiguities. The prior/posterior block (i.e., Pr/Po block in Fig. 1) is designed to build the distribution of enhancement. We note that both Pr and Po need to construct a mean and a standard deviation distributions. After that, embeddings sampled form Pr/Po block are input to the AdaIN to transform the feature statistics. Let ${\bm{f}} \in {\mathbbm{R}^{B \times C \times H \times W}}$ refers to the data matrix of Pr/Po block’s input, where $B$, $C$, $H$, $W$ indicate the batch size, number of channels, the height, and the width, respectively. First, we calculating the mean and standard deviation of each channel of $\bm{f}$. Then, we adopt ${1 \times 1}$ convolutions to obtain ${\bm{\mu} \in {\mathbbm{R}^{B \times N \times 1 \times 1}}}$ and ${\bm{\sigma} \in {\mathbbm{R}^{B \times N \times 1 \times 1}}}$ from the mean vector. Similarly, we adopt ${1 \times 1}$ convolutions to obtain ${\bm{m} \in {\mathbbm{R}^{B \times N \times 1 \times 1}}}$ and ${\bm{v} \in {\mathbbm{R}^{B \times N \times 1 \times 1}}}$ from the standard deviation vector. Finally, ${\bm{\mu}}$ and ${\bm{\sigma}}$ are applied to build the $N$-dimensional Gaussian distribution of the mean ($\mathcal{N}_{\rm{m}}$). ${\bm{m}}$ and ${\bm{v}}$ are applied to build the $N$-dimensional Gaussian distribution of the standard deviation ($\mathcal{N}_{\rm{s}}$). As the two distributions are constructed, we extract random samples from them, which can be expressed as:
\begin{equation}
{\bm{a}} \sim \mathcal{N}_{\rm{m}}\left( {\bm{\mu}\left( {{\bf{x}}} \right),\bm{\sigma}^2\left( {{\bf{x}}} \right)} \right)
\end{equation}
\vspace{-4mm}
\begin{equation}
{\bm{b}} \sim \mathcal{N}_{\rm{s}}\left( {\bm{m}\left( {{\bf{x}}} \right),\bm{v}^2\left( {{\bf{x}}} \right)} \right)
\end{equation}
where $\bm{a}$ and ${\bm{b}}$ are two random samples from the mean and standard deviation distributions, respectively. Note that, in the testing phase, the latent codes ($\bm{a}$ and $\bm{b}$) are only dependent on the input image $\bf{x}$. While in the training phase we leverage input image $\bf{x}$ and corresponding reference image $\bf{y}$ to learn the posterior distributions, which will be described in the next subsection.

Random samples ${\bm{a}}$ and ${\bm{b}}$ are further injected into the AdaIN module to transform the statistics of received features. Typical AdaIN receives a content input and a style input, and simply aligns the mean and standard deviation of the content input to match those of the style input across channels. For UIE, the style input is unavailable in advance. Therefore, we propose to align the mean and standard deviation of received features based on random activations extracted from the learned distributions.  

\subsection{Training and Testing}
PUIE-Net is trained following the standard training procedure of CVAE, i.e., by minimizing the variational lower bound (Eq. 12). The main difference with respect to training a deterministic enhancement model is that our training process additionally needs to find a meaningful embedding of the enhancement statistics in the latent space. This is addressed by introducing a posterior network (i.e., the bottom branch in Fig. 1), that learns to recognize the posterior features, and map those to two posterior distributions of the mean and standard deviation. Samples from the posterior distributions can be formalized as:
\begin{equation}
{\bm{a}} \sim \mathcal{N}_{\rm{m}}\left( {\bm{\mu} \left( {{\bm{y}},{\bf{x}}} \right),\bm{\sigma}^2 \left( {{\bm{y}},{\bf{x}}} \right)} \right)
\end{equation}
\vspace{-4mm}
\begin{equation}
{\bm{b}} \sim \mathcal{N}_{\rm{s}}\left( {\bm{m} \left( {{\bm{y}},{\bf{x}}} \right),\bm{v}^2 \left( {{\bm{y}},{\bf{x}}} \right)} \right)
\end{equation}
where $\bm{a}$ and ${\bm{b}}$ are two random samples from the mean and standard deviation posterior distributions, respectively. During the training, random samples $\bm{a}$ and ${\bm{b}}$ are fed into the AdaIN module to predict the enhanced image. The enhancement loss $L_{\rm{e}}$ penalizes the differences between the output of PUIE-Net and the reference. $L_{\rm{e}}$ is formulated as:
\begin{equation}
{L_{\rm{e}}} = {L_{\rm{mse}}} + \lambda {L_{\rm{vgg16}}}
\end{equation}
where ${L_{\rm{mse}}}$ denotes the mean square error loss and ${L_{\rm{vgg16}}}$ denotes the perceptual loss explored by \cite{56}, $\lambda$ refers to the weight.

Apart from minimizing the enhancement loss, Kullback-Leibler (KL) divergences are employed to assimilate the posterior distributions and the prior distributions:
\begin{equation}
L_m = {D_{{\rm{KL}}}}\left( {{\mathcal{N}_{\rm{m}}}\left( {{\bf{x}}} \right)\left\| {{\mathcal{N}_{\rm{m}}}\left( {{\bm{y}},{\bf{x}}} \right)} \right.} \right)
\end{equation}
\vspace{-4mm}
\begin{equation}
L_s = {D_{{\rm{KL}}}}\left( {{\mathcal{N}_{\rm{s}}}\left( {{\bf{x}}} \right)\left\| {{\mathcal{N}_{\rm{s}}}\left( {{\bm{y}},{\bf{x}}} \right)} \right.} \right)
\end{equation}
where $D_{\rm{KL}}$ refers to the KL divergence between two distributions. Finally, the total loss function for training PUIE-Net is the weighted sum of 
above losses:
\begin{equation}
L = {L_e} + \beta (L_m + L_s)
\end{equation}
where $\beta$ is the weight. In the testing phase, we apply the network ${n}$ times to the same input image to predict ${n}$ enhancement variants. Note that only PAdaIN and the output block need to be re-evaluated. Diverse enhancement solutions provide users with multiple alternative results for display or analysis. More importantly, a set of samples provide sufficient inferring data for the consensus process. In this paper, the default consensus processes are the Monte Carlo likelihood estimation (MC) and Maximum Probability estimation (MP). MC predicts a final result via averaging a group of possible samples. MP takes the enhancement sample with the maximum probability as the final estimation. Equations (1) and (2) in section 3.1 describe the formulation of MC and MP, receptively.

\subsection{Training Data Generation}

\begin{figure}[!t]
	\centerline{\includegraphics[scale=0.48]{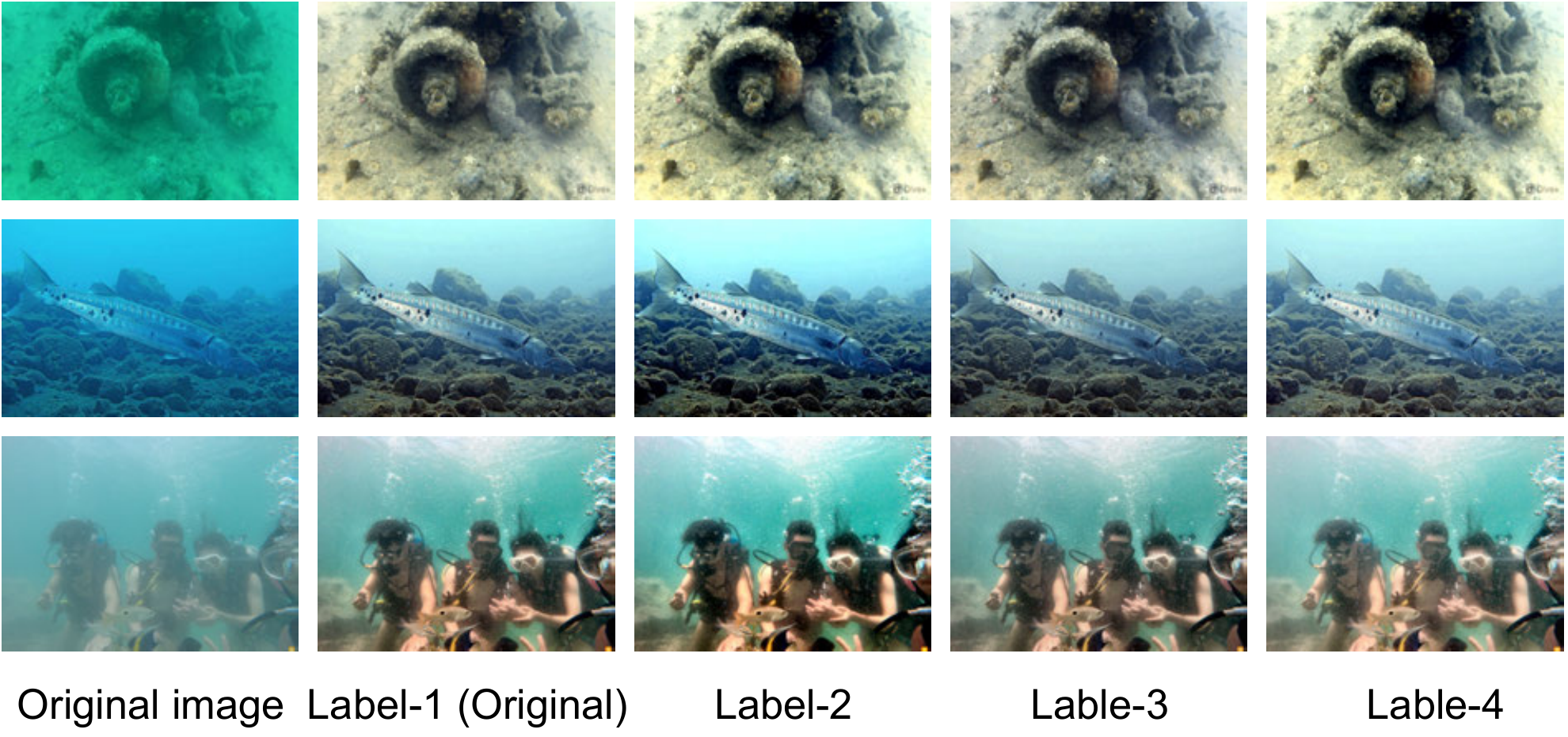}}
	\caption{Examples of our new UIE dataset. Label 2-4 denote the new labels generated by contrast adjustment, saturation adjustment, and gamma correction, respectively.}
\end{figure}

One obstacle before training our probabilistic network is that the existing UIE dataset generally only provides a single reference map for each degraded underwater image. To apply the probabilistic network, we re-build the existing UIE dataset by generating multiple reference images. Our new dataset is based on UIEBD~\cite{1}, a real-world UIE dataset that contains 890 underwater images and corresponding reference maps. In the original UIEBD, the authors utilize 12 state-of-the-art enhancement algorithms to generate the potential ground truth. With raw underwater images and the 12 enhanced results, the authors invite volunteers to perform pairwise comparisons and subjectively select the best one among twelve candidates as the final reference image. Based on UIEBD, we create ambiguities by performing contrast and saturation adjustment, and gamma correction. We adopt these methods because the distortions of underwater images are mainly reflected in contrast, saturation, brightness, and colors. Note that we aim to generate ambiguous labels rather than significantly alter the original label. The contrast and saturation adjustment are performed via a simple linear transformation $y = (x - m) \times \alpha + x$, where $x$ and $y$ refer to the input and output, $m$ denotes the mean of each channel. $\alpha$ is the adjustment coefficient. For contrast adjustment, $\alpha$ is the same for all pixels. For saturation adjustment, $\alpha$ is determined by each pixel itself. To generate a more reliable reference map, we first create two adjusted versions (i.e., over and under adjustment) per method and then choose the better one as the potential label. As a result, we obtain four reference maps (including the original label) for each raw underwater image, which can reflect the uncertainty during the ground truth recording. We show an example of the new dataset in Fig. 2.

\section{Experiments}

In this section, we will first show the detailed experimental settings including training and test datasets, performance criteria, compared methods, and the implementation details. Then, we quantitatively and qualitatively evaluate our method against several state-of-the-art algorithms on two UIE datasets. 

\subsection{Experiment Settings}

\textbf {Training and testing datasets:} Two UIE datasets are used for performance verification. The first one is the new UIEBD dataset that is built in this paper. As described in section 3.4, we create multiple labels for each raw underwater image. We use the first 700 original images and corresponding reference images for training, and the rest for testing. The second dataset is RUIE~\cite{3}, which only contains raw underwater images. RUIE is a large-scale underwater dataset that contains three subsets, including an underwater image quality subset (3630 images), an underwater color cast subset (300 images), and an underwater higher-level task-driven subset (300 images). In this paper, we use the underwater image quality subset for testing since it contains different levels of image quality and various underwater scenes. Note that our model is trained on UIEBD and tested on both UIEBD and RUIE.

\noindent\textbf{Performance criteria:} In the case of applying a probabilistic network for UIE, we not only want to compare a deterministic estimation with a unique reference image, but also we are interested in the distributions of enhancement. To analyze the learned distributions, we perform subjective comparisons by visualizing the latent space. To evaluate the enhancement performance, we adopt SSIM~\cite{57}, PSNR, DeltaE (CIE2000 standard)~\cite{63}, and NIQE~\cite{60} to measure the image quality objectively. SSIM, PSNR, and DeltaE are full-reference metrics calculated based on the original label in UIEBD for a fair comparison with existing methods. Note that DeltaE is used for color difference evaluation. NIQE is a no-reference metric and it does not need reference images. Additionally, we conduct subjective tests to understand how users prefer the results generated by each UIE method. We use Mean Opinion Score (MOS) to quantify the subjective evaluation. 20 participants (10 male and 10 female) are invited to join the subjective test. Raw and enhanced underwater images are simultaneously displayed on a screen. The subjective score of each image is rated on a five-level scale: 5 (excellent), 4 (good), 3 (fair), 2 (poor), and 1 (bad), according to the following measures: color distortion, contrast enhancement, naturalness preservation, brightness improvement, and artifacts. We note that since RUIE only contains raw underwater images, we adopt NIQE and MOS to measure the model performance. 

\noindent\textbf{Compared methods:} We compare PUIE-Net with 9 UIE methods, including three model-free methods (GC, Retinex~\cite{13} and Fusion~\cite{11}), three prior-based methods (IBLA~\cite{25}, Histogram-Prior~\cite{20} and Haze-line~\cite{61}), and three deep learning-based approaches (Water-Net~\cite{1}, Ucolor~\cite{33}, and LC-Net~\cite{62}). We record the results of all competitors by conducting the same experiments using the original implementations provided by the authors for comparison fairness.

\noindent\textbf{Implementation details:} PUIE-Net is implemented in the Pytorch framework and trained on an NVIDIA RTX 2080Ti GPU with ADAM optimizer. The learning rate is $1 \times 10^{-4}$, the batch size is 4, and the patch size is $256 \times 256$. We augment the training data with rotation, flipping horizontally and vertically to promote network generalization. The dimension $N$ of the latent space is 20. We adopt ${1 \times 1}$ convolutions to broadcast the samples to the desired number of channels before input to AdaIN. The parameter $\lambda$ and $\beta$ are empirically set as 1. The default kernel size of convolution layers is ${3 \times 3}$ and the number of channels is 64. The default sampling times for calculating MC estimations are 20.

\begin{figure}[!t]
	\centerline{\includegraphics[scale=0.32]{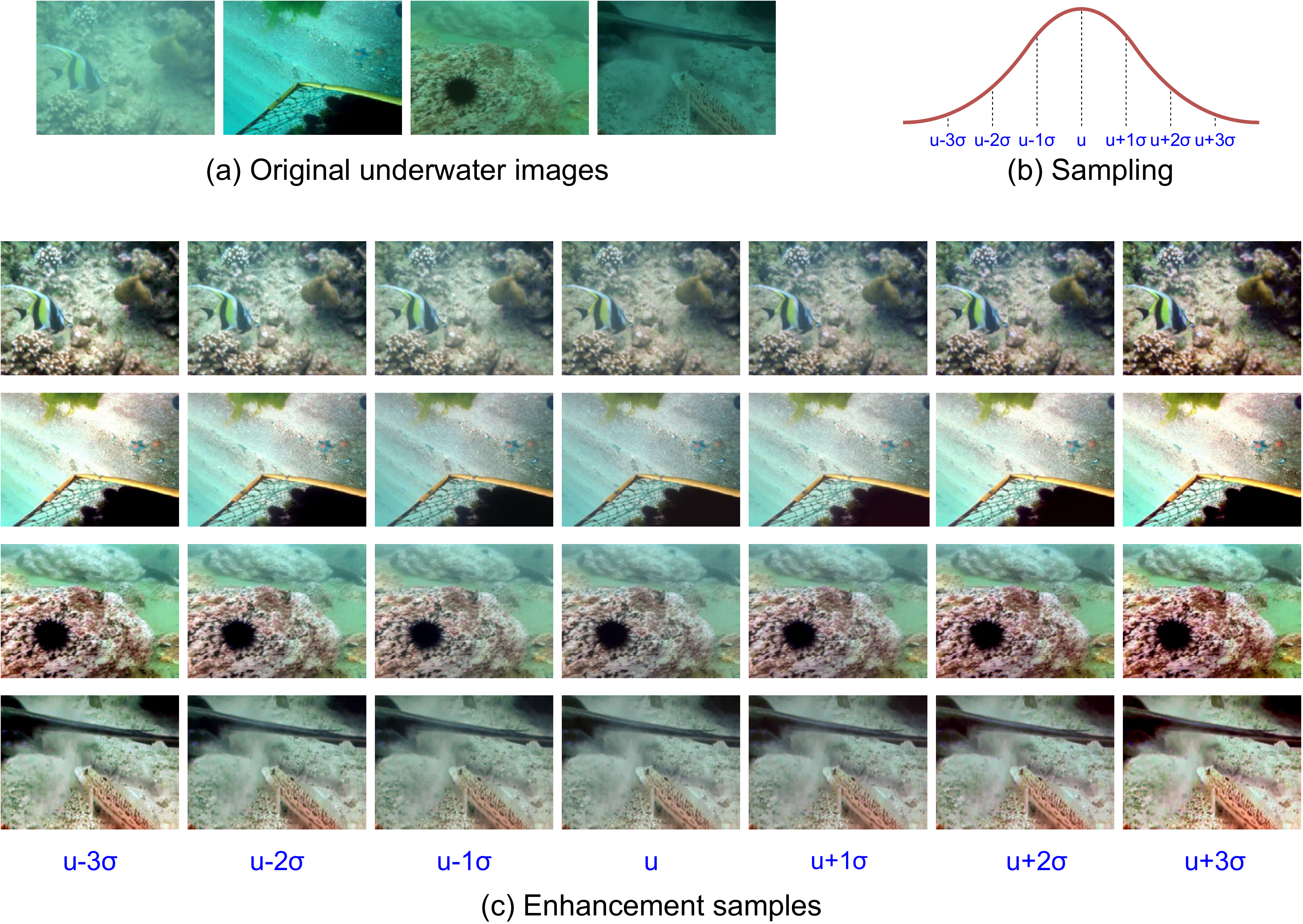}}
	\caption{Visualization of the enhancement distribution of UIE. We show the original image in (a). The sampling scheme used for visualization is presented in (b) and the enhancement samples are shown in (c). $\mu$ and $\sigma$ in (b) denote the mean and standard deviation of the distribution, respectively.}
\end{figure}

\subsection{Analysis of Enhancement Distribution}

Although a lot of UIE algorithms have been developed, PUIE-Net is the first method that learns the distribution of enhancement and explicitly takes multi-solution into account. Fig. 3 shows the original image and PUIE-Net samples. We manually control the sampling interval for better visualization. From Fig. 3, we can make the following observations: 1) Each PUIE-Net sample has a different but reasonable appearance. Enhanced samples with higher sampling probability show relatively mild enhancement. Enhanced samples with lower sampling probability show strong contrast and color adjustment. This demonstrates that our method can learn the meaningful distribution of UIE and can generate diverse enhancement predictions. 2) A set of enhanced samples not only provide multiple alternatives for display or analysis but also offer effective data for the consensus process to further obtain a robust and stable result. Compared with deterministic learning-based methods which have to make a compromise between possible results, our method is more flexible and can reduce the influence of biased labels in the existing UIE dataset.

\subsection {Performance Comparison}

\begin{table}[!t]
	\caption{Quantitative comparison on UIEBD and RUIE datasets. The best result is highlighted in red and the second best one is in blue.}
	\label{tab1}
	\renewcommand\tabcolsep{2pt}	
	\centering
	\begin{tabular}{c|ccccc|cc}
	\hline
	\multirow{2}*{Method} & \multicolumn{5}{|c|}{UIEBD}  & \multicolumn{2}{|c}{RUIE} \\
	\cline{2-8}
	~ & PSNR $\uparrow$ & SSIM $\uparrow$ & DeltaE $\downarrow$ & NIQE $\downarrow$ & MOS $\uparrow$ & NIQE $\downarrow$ & MOS $\uparrow$ \\
	\hline
	GC             &16.14 &0.761 &16.11 &3.789  &2.6   &4.656  &2.4\\
	Retinex~\cite{13}        &17.53 &0.773 &14.82 &4.074 &2.7   &4.593  &2.5\\
	Fusion~\cite{11}         &21.18 &0.822 &\textcolor{blue}{9.079} &3.747 &3.4    &\textcolor{blue}{4.488}  &2.9\\
	IBLA~\cite{25}           &18.51 &0.762 &18.64 &4.290 &2.2   &4.767  &1.9\\
	Histogram-Prior~\cite{20} &14.39 &0.573 &15.69   &3.780 &2.4  &\textcolor{red}{4.486}  &2.5\\
	Water-Net~\cite{1}      &19.31 &0.830 &10.14 &3.879 &3.3    &4.491  &3.3\\
	Haze-line~\cite{61}  &14.97 &0.669 &17.23  &3.830 &2.4  &4.620 &1.8\\
	Ucolor~\cite{33}         &\textcolor{blue}{21.65} &0.840 &\textcolor{red}{8.646} &3.786 &3.7  &4.755  &3.0\\
	LC-Net~\cite{62} &18.54 &0.812 &14.25 &3.800 &3.3  &4.721 &2.9\\
	PUIE-Net (MC)  &\textcolor{red}{21.86} &\textcolor{red}{0.870} &9.556 &\textcolor{red}{3.626} &\textcolor{red}{4.2}  &4.512  &\textcolor{red}{3.7}\\ 
	PUIE-Net (MP) &21.05 &\textcolor{blue}{0.854} &10.26 &\textcolor{blue}{3.668} &\textcolor{blue}{4.1}  &4.555  &\textcolor{blue}{3.6}\\ 		
	\hline
	\end{tabular}
\end{table}

Tab. 1 presents the quantitative comparisons on UIEBD and RUIE datasets. From Tab. 1, we can perceive that PUIE-Net achieves favorable performance and outperforms other methods. Specifically, prior-based approaches obtain relatively poor results since this kind of method is highly dependent on the used prior knowledge and the predefined imaging model. We found that the performance of MC is significantly better than the others. This is reasonable since averaging a set of samples can reduce enhancement bias in the testing phase. We further present visual comparisons in Fig. 4. As can be seen, although most of the methods can somehow enhance the contrast, severe visual defects caused by the unsatisfactory adjustment of colors or artifacts remain. For example, GC and Retinex show unnatural colors and saturation, smearing image details. Prior-based methods can improve the contrasts, however, the colors are seriously degraded in these cases. Water-Net and Ucolor tend to generate over/under-enhanced results. Our method works well on all these cases and the result looks more clean and natural with fine-grained textures. 

\begin{figure}[!t]
	\centerline{\includegraphics[scale=0.225]{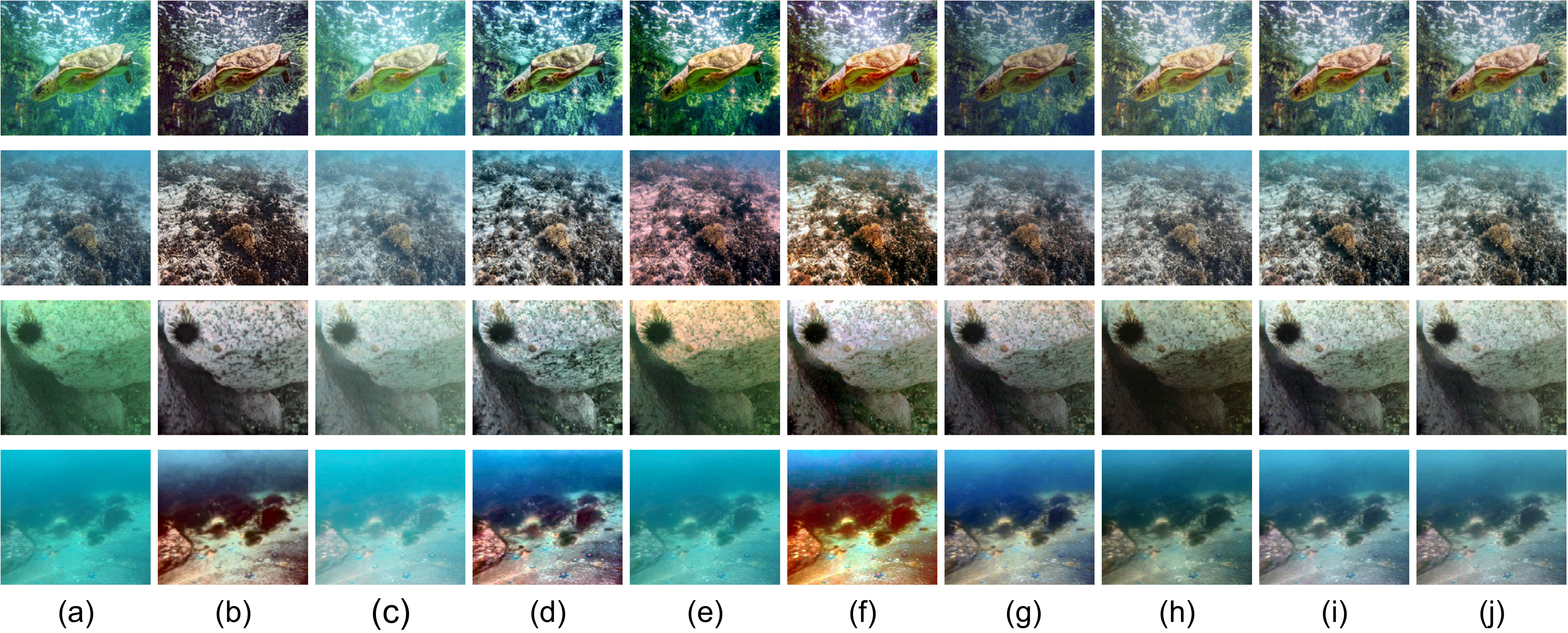}}
	\caption{Comparisons of visual results on UIEBD and RUIE datasets. (a) Original image. (b) Retinex. (c) GC. (d) Fusion. (e) IBLA. (f) Histogram-Prior. (g) Water-Net. (h) Ucolor. (i) PUIE-Net (MC). (j) PUIE-Net (MP).}
\end{figure}

\subsection{Impact of Sampling Times}

Since we apply the consensus process to obtain a deterministic result, it is necessary to analyze the influence of sampling times. We calculate the standard deviation of PSNR and SSIM under different sampling times on the UIEBD dataset. Note that we run the model 10 times to calculate the standard deviation of PSNR and SSIM (i.e., std-PSNR and std-SSIM) for each raw underwater image. The mean std-PSNR and std-SSIM of all test images are listed in Tab. 2. As can be observed, as the number of samples increases, the standard deviation of PSNR and SSIM first reduces and then becomes stable after 20 sampling times. This demonstrates that PUIE-Net can capture the diversity of enhanced images, and the consensus process can estimate a stable result based on multiple predictions. Note that increasing sampling times can further improve the stability of final predictions. In this paper, we set the default sampling times of MC estimation as 20 to better balance the stability and running time.

\begin{table}[!t]
	\caption{Impact of sampling times.}
	\label{tab2}
	\renewcommand\tabcolsep{4pt}	
	\centering
	\begin{tabular}{c|cccccccc}
		\hline	
		Sampling times   & 1  & 2 & 4 & 6 & 8 & 10 & 20 & 50\\
		\hline
		Mean std-PSNR    & 0.726 & 0.512 & 0.380 &  0.326 &  0.292 & 0.253 & 0.175 & 0.166\\
		Mean std-SSIM    & 0.008 & 0.004 & 0.003 &  0.003 &  0.003 & 0.002 & 0.002 & 0.001\\
		\hline
	\end{tabular}
\end{table}

\subsection{Discussion}

\subsubsection{Consensus Process}
The key idea of PUIE-Net is to learn the enhancement distribution and employ a consensus process to get the final prediction. Therefore, the effectiveness of the consensus process has a significant impact on the final results. In this paper, we have designed two consensus processes i.e., MC and MP. The former estimates an enhanced image by averaging a group of samples while the latter takes the image with the highest sampling probability as the final result. We consider that many other approaches can be applied to get the final result. For example, one can perform an image quality assessment method to select the best result according to the visual quality. In this case, both subjective and objective metrics can be used and the enhancement performance is highly dependent on the used evaluation method. Quality based selection can be expressed as:
\begin{equation}
p\left( {y\left| {\bf{x}} \right.} \right) \approx \mathop {\max }\limits_s Q\left( {p\left( {{\bf{y}}\left| {{{\bf{z}}^{\left( s \right)}},{\bf{x}}} \right.} \right),{{\bf{z}}^{\left( s \right)}} \sim p\left( {{\bf{z}}\left| {\bf{x}} \right.} \right)} \right)
\end{equation}
where $Q$ denotes the subjective/objective image quality assessment function.

\subsubsection{Relationship with Unsupervised Method}
The degradation of underwater images is non-reversible and complex, it is impracticable to record the ground truth for training a supervised UIE model directly. Therefore, we consider that unsupervised methods are more suitable for addressing the UIE problem. In this paper, instead of concentrating on developing an unsupervised method, we aim at making full use of the biased labels in existing UIE datasets. Biased reference maps of real-world underwater images are possible and cheap to obtain~\cite{1,33}. With such reference maps, we proposed to leverage an AadIN module and a conditional variational autoencoder to learn the enhancement distribution. Therefore, the goal of both PUIE-Net and unsupervised UIE methods is to tackle the challenge of learning a UIE model without the ground truth. But our framework provides a new means to address the UIE problem i.e., resolving UIE into distribution estimation and consensus process based on biased reference maps, which is different from the unsupervised methods. 

\section{Conclusion}

In this paper, we introduce PUIE-Net, a novel probabilistic network for underwater image enhancement. Specifically, we propose to learn the distribution of enhanced images rather than directly estimate a single result with certainty. This allows us to handle the label ambiguity issue of underwater image enhancement. The main idea is to leverage random style attributes from two posterior distributions constructed by conditional variational autoencoders, to transform the global enhancement statistics of input features. Moreover, with the consensus process, final enhancement results can be inferred via integrating a set of predictions. We demonstrate that PUIE-Net can effectively produce a set of reasonable results and the visual quality of the consensus estimation is highly competitive on two real-world UIE datasets. In the future, we plan to extend our method to other image enhancement tasks such as low-light image enhancement, dehazing, and denoising, where the ground truth image is difficult to obtain.

\section*{Acknowledgements}
This study was partially supported by National Natural Science Foundation of China under Grants 82172033, 61971369, U19B2031, Science and Technology Key Project of Fujian Province 2019HZ020009, Fundamental Research Funds for the Central Universities 20720200003, Tencent Open Fund, and the State Scholarship Fund.

\clearpage
%
%
\bibliographystyle{splncs04}
\bibliography{cas-refs}
\end{document}